\documentclass[11pt,a4paper]{article}
\usepackage[hyperref]{naaclhlt2019} 
\usepackage{times}
\usepackage{latexsym}

\usepackage{graphicx}
\usepackage{ctable}
\usepackage{multirow}
\usepackage{amsmath,amsbsy,amsthm,amssymb}

\newtheorem{theorem}{Theorem}%[section]

\newcommand{\E}{\operatorname*{\mathbb{E}}}

\newcommand\jj{^{{(j)}}}
\newcommand{\mmd}{\operatorname{MMD}}
\newcommand{\diag}{\operatorname{diag}}

\usepackage{bm}

\usepackage{url} 
\renewcommand{\toprule}{\hline}
\renewcommand{\midrule}{\hline}
\renewcommand{\bottomrule}{\hline}
\aclfinalcopy % Uncomment this line for the final submission

\usepackage{fancyhdr}
\title{Stochastic Wasserstein Autoencoder for Probabilistic Sentence Generation}

\author{Hareesh Bahuleyan,$^\dag$ Lili Mou,$^\dag$ Hao Zhou,$^\ddag$ Olga Vechtomova$^\dag$\\
$^\dag$University of Waterloo\quad $^\ddag$ByteDance AI Lab\\
\url{hpallika@uwaterloo.ca}\quad \url{doublepower.mou@gmail.com}\\
\url{zhouhao.nlp@bytedance.com}\quad \url{ovechtomova@uwaterloo.ca}}

\date{}

\newcommand{\n}{^{(n)}}
\newcommand{\m}{^{(m)}}
\newcommand{\stdnorm}{\mathcal{N}(\bm 0, \mathbf I)}
\newcommand{\KL}{\operatorname{KL}}

\begin{document}
\maketitle

\begin{abstract}
The variational autoencoder (VAE) imposes a probabilistic distribution (typically Gaussian) on the latent space and penalizes the Kullback--Leibler (KL) divergence between the posterior and prior. In NLP, VAEs are extremely difficult to train due to the problem of KL collapsing to zero. One has to implement various heuristics such as KL weight annealing and word dropout in a carefully engineered manner to successfully train a VAE for text. In this paper, we propose to use the Wasserstein autoencoder (WAE) for probabilistic sentence generation, where the encoder could be either stochastic or deterministic. We show theoretically and empirically that, in the original WAE, the stochastically encoded Gaussian distribution tends to become a Dirac-delta function, and we propose a variant of WAE that encourages the stochasticity of the encoder. Experimental results show that the latent space learned by WAE exhibits properties of continuity and smoothness as in VAEs, while simultaneously achieving much higher BLEU scores for sentence reconstruction.\footnote{Our code is availabe at \url{https://github.com/HareeshBahuleyan/probabilistic_nlg}\\ A preliminary version of this paper was preprinted at \url{https://arxiv.org/abs/1806.08462}}
\end{abstract}

\section{Introduction}

\thispagestyle{fancy}
Natural language sentence generation in the deep learning regime typically uses a recurrent neural network (RNN) to predict the most probable next word given previous words~\cite{mikolov2010recurrent}. Such RNN architecture can be further conditioned on some source information, for example, an input sentence, resulting in a sequence-to-sequence (Seq2Seq) model.

Traditionally, sentence generation is accomplished in a \textit{deterministic} fashion, i.e., the model uses a deterministic neural network to encode an input sentence to some hidden representations, from which it then decodes an output sentence using another deterministic neural network.

\newcite{bowman2015generating} propose to use the variational autoencoder \cite[VAE,][]{kingma2013auto} to map an input sentence to a probabilistic continuous latent space. VAE makes it possible to generate sentences from a distribution, which is desired in various applications. For example, in an open-domain dialog system, the information of an utterance and its response is not necessarily a one-to-one mapping, and multiple plausible responses could be suitable for a given input. Probabilistic sentence generation makes the dialog system more diversified and more meaningful~\cite{HVED,VarAtt}. Besides, probabilistic modeling of the hidden representations serves as a way of posterior regularization~\cite{VNMT}, facilitating interpolation~\cite{bowman2015generating} and manipulation of the latent representation~\cite{control}.

However, training VAEs in NLP is more difficult than the image domain~\cite{kingma2013auto}.  
The VAE training involves a reconstruction loss and a Kullback--Leibler (KL) divergence between  the posterior and prior of the latent space.  In NLP, the KL term tends to vanish to zero during training, leading to an ineffective latent space. Previous work has proposed various engineering tricks  to alleviate this problem, including KL annealing and word dropout~\cite{bowman2015generating}.

In this paper, we address the difficulty of training VAE sentence generators by using a Wasserstein autoencoder \cite[WAE,][]{wae}. WAE modifies VAE in that it requires the integration of the posterior to be close to its prior, where the closeness is measured with empirical samples drawn from the distributions. In this way, the encoder could be either stochastic or deterministic, but the model still retains probabilistic properties. 

Moreover, we show both theoretically and empirically that the stochastic Gaussian encoder in the original form tends to be a Dirac-delta function. We thus propose a WAE variant that encourages the encoder's stochasticity by penalizing an auxiliary KL term.

Experiments show that the sentences generated by WAE exhibit properties of continuity and smoothness as in VAE, while achieving a much higher reconstruction performance. Our proposed variant further encourages the stochasticity of the encoder.  More importantly, WAE is robust to hyperparameters and much easier to train, without the need for KL annealing or word dropout as in VAE. In a dialog system, we demonstrate that WAEs are capable of generating better quality and more diverse sentences than VAE.

\begin{figure}[!t]
\includegraphics[width=\linewidth]{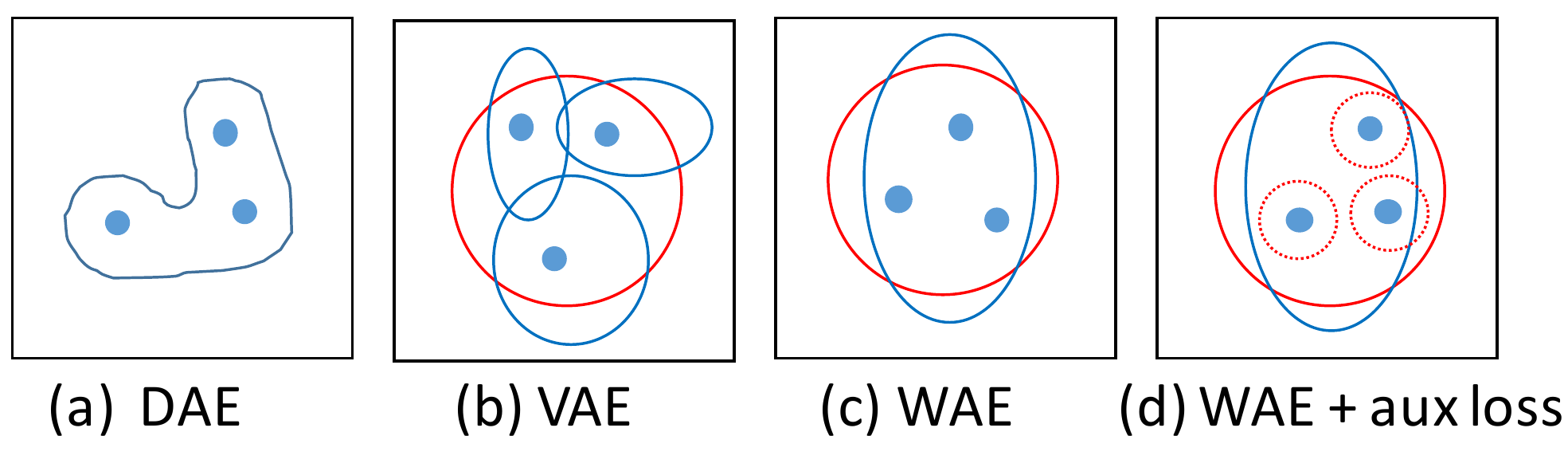}
\caption{The latent space of the deterministic autoencoder (DAE), variational autoencoder (VAE), Wasserstein autoencoder (WAE), as well as WAE with by our KL penalty. {\color{blue}Blue} circles: Posterior or aggregated posterior distributions of data in the latent space. {\color{red}Red} circles: Regularizations of the posterior.}\label{fig:diagram}
\end{figure}

\section{Probabilistic Sentence Generation}

\paragraph{Base Model: Deterministic Autoencoder (DAE).}

DAE encodes an input sentence with a recurrent neural network (RNN) and then decodes the same sentence through another RNN.

For the encoder, the hidden state of the last word is represented as the latent space of the input sentence $\mathbf x$. The latent representation is denoted as $\bm z$. We feed $\bm z$ to the decoder RNN, which predicts one word at a time using a softmax layer, given by $p(\mathrm{x}_t|\bm 
z, \mathbf x_{< t})$.%=\operatorname{softmax}(W\bm h_t\de+\bm b)$, where $W$ and $\bm b$ are parameters, and $\bm h_t\de$ is the hidden state at time step $t$ of the decoder. 
The training objective for DAE is the sequence-aggregated cross-entropy loss, given by
\begin{equation}
J=- \sum_{n=1}^N\sum_{t=1}^{|\mathbf x\n|} \log p(\mathrm{x}_t\n|\bm 
z\n, \mathbf x_{<t}\n)
\label{eqn:Jxent}
\end{equation}
where superscript $(n)$ indicates the $n$th data point among $1,\cdots, N$.

In DAE, the latent space is encoded and then decoded in a deterministic way, i.e., there is no probabilistic modeling of the hidden space. The hidden representations of data  may be located on an arbitrary manifold (Figure~\ref{fig:diagram}a), which is not suitable for probabilistic generation.

\paragraph{Variational Autoencoder (VAE).}
VAE extends DAE by imposing a prior distribution $p(\bm z)$ on the latent variable $\bm z$, which is typically set to the standard normal $\stdnorm$~\cite{kingma2013auto}. Given an input sentence $\mathbf x$, we would like to model the posterior of $\bm z$ by another normal distribution, $q(\bm z|\mathbf x)=\mathcal N(\bm\mu_\text{post}, \diag \bm\sigma_\text{post}^2)$, where $\bm\mu_\text{post}$ and $\bm\sigma_\text{post}^2$ are the outputs of the encoder.

In the training of VAE, $\bm z$ is sampled from $q(\bm z|\mathbf x)$, and the training objective is to maximize a variational lower bound of the likelihood of data. This is equivalent to minimizing the (expected) reconstruction loss similar to (\ref{eqn:Jxent}), while being regularized by the KL divergence between $q(\bm z|\bm x)$ and $p(\bm z)$, given by
\begin{align}\nonumber
J= &\sum_{n=1}^N\Big[ - \underset{\bm z\n\sim q}{\mathbb E} \sum_{t=1}^{|\mathbf x\n|}\log p(\mathrm{x}_t\n|\bm 
z\n, \mathbf x_{<t}\n)\\
&+\lambda_\text{VAE}\cdot\KL(q(\bm z\n|\mathbf x\n)\| p(\bm z))\Big]
\label{eqn:Jvae}
\end{align}
where in the expectation $\bm z\n$ is sampled from $q(\bm z|\mathbf x\n)$ and $\lambda_\text{VAE}$ is a hyperparameter balancing the two terms.

Since VAE penalizes the divergence of $\bm z$'s posterior from its prior, it serves as a way of posterior regularization, making it possible to generate sentences from the continuous latent space.

However, the two objectives in~(\ref{eqn:Jvae}) are contradictory to each other, as argued by~\newcite{wae}. VAE pushes the posterior of $\bm z$, given any input $\mathbf x\n$, to be close to its prior, i.e., every blue ellipse in Figure~\ref{fig:diagram}b should be close to the red one. This makes perfect reconstruction impossible.

Further, VAE is difficult to train in NLP due to the problem of KL collapse, where the KL term tends to be zero, meaning that the encoder captures no information and the decoder learns an unconditioned language model. This phenomenon is observed in variational auto-regressive decoders using RNN.
To alleviate this problem, existing tricks include KL annealing and word dropout~\cite{bowman2015generating}, but both require extensive engineering.

\paragraph{Wasserstein Autoencoder (WAE).}
 An alternative way of posterior regularization is to impose a constraint that the \textit{aggregated posterior} of $\bm z$ should be the same as its prior~\cite{wae}, i.e., $q(\bm z) \overset{\text{def}}= \sum_\mathbf{x} q(\bm z|\mathbf x)p_\mathcal{D}(\mathbf x)\overset{\text{set}}=p(\bm z)$, where $p_\mathcal{D}$ is the data distribution. This is also demonstrated in Figure~\ref{fig:diagram}c. By contrast, VAE requires that $q(\bm z|\mathbf x)$ should be close to $p(\bm z)$ for every input sentence $\mathbf{x}$.

For computational purposes, \newcite{wae} relax the above constraint by penalizing the Wasserstein distance between $q(\bm z)$ and $p(\bm z)$. In particular, it is computed by the Maximum Mean Discrepancy (MMD), defined as
\begin{equation}\nonumber 
\mmd = \left\lVert\int k(\bm z, \cdot)\operatorname d\!P(\bm z) - \int k(\bm z, \cdot)\operatorname d\!Q(\bm z) \right\rVert_{\mathcal H_k}
\label{eqn:Wdist}
\end{equation}
where $P(\bm z)$ and $Q(\bm z)$ are cumulative density functions. $\mathcal H_k$ refers to the reproducing kernel Hilbert space defined by the kernel $k$, which is often chosen as the inverse multiquadratic kernel $k(x,y) = \frac{C}{C + \left\lVert x-y \right\rVert_2^2}$ for high-dimensional Gaussians.

One advantage of the Wasserstein distance is that it can be estimated by empirical samples as

\begin{small}
	\begin{align}
	&\widehat{\mmd}= \frac{1}{N(N-1)}\sum_{n \neq m}k(\bm z\n, \bm z\m) \\ \nonumber
	&+ \frac{1}{N(N-1)}\sum_{n \neq m}k(\widetilde{\bm z}\n, \widetilde{\bm z}\m) - \frac{1}{N^2}\sum_{n,m}k(\bm z\n, \widetilde{\bm z}\m)
	\label{eqn:mmd}
	\end{align}
\end{small}

\noindent\!\,where  $\widetilde{\bm z}\n$ is a sample from the prior $p(\bm z)$, and ${\bm z}\n$ is a sample from the aggregated posterior $q(\bm z)$, which is obtained by sampling $\mathbf x\n$ from the data distribution and then sampling ${\bm z}\n$ from $q(\bm z|\mathbf x\n)$. In summary, the training objective of WAE is

\vspace{-.3cm}
\begin{small}
	\begin{equation}
	J_\text{WAE}=- \sum_{n=1}^N\sum_{t=1}^{|\mathbf x\n|} \log p(\mathrm{x}_t\n|
	{\bm z}\n, \mathbf x_{<t}\n) + \lambda_\text{WAE}\cdot\widehat{\mmd}
	\label{eqn:Jwae}
	\end{equation}
\end{small}
\noindent\!\!where $\lambda_\text{WAE}$ balances the MMD penalty and the reconstruction loss.

Alternatively, the dual form (adversarial loss) can also be used for WAE~\cite{ARAE}. In our preliminary experiments, we found MMD similar to but slightly better than the adversarial loss.  The difference between our work and  \newcite{ARAE}---who extend the original WAE to sequence generation---is that we address the KL annealing problem of VAE and further analyze the stochasticity of WAE from a theoretical perspective, as follows.

\paragraph{WAE with Auxiliary Loss.} In WAE, the aggregated posterior $q(\bm z)$ involves an integration of data distribution, which allows using a deterministic function to encode $\bm z$ as $\bm z=\bm f_\text{encode}(\mathbf x)$ as suggested by \newcite{wae}. This would largely alleviate the training difficulties as in VAE, because backpropagating gradient into the encoder no longer involves a stochastic layer.

The stochasticity of the encoder, however, is still a desired property in some applications, for example, generating diverse responses in a dialog system.
We show both theoretically and empirically that a \textit{dangling} Gaussian stochastic encoder could possibly degrade to a deterministic one.
\begin{theorem}
Suppose we have a Gaussian family $\mathcal{N}(\bm\mu,\diag\bm\sigma^2)$, where  $\bm \mu$ and $\bm \sigma$ are parameters. The covariance is diagonal, meaning that the variables are independent. If the gradient of $\bm \sigma$ completely comes from sample gradient and $\bm \sigma$ is small at the beginning of training, then the Gaussian converges to a Dirac delta function with stochastic gradient descent, i.e., $\bm \sigma\rightarrow\bm 0$. (See Appendix~\ref{sec:A} for the proof.)\qed
\end{theorem} 

To alleviate this problem, we propose a simple heuristic that encourages the stochasticity of the encoder. In particular, we penalize, for every data point, a KL term between the predicted posterior $q(\bm z|\bm x)=\mathcal N(\bm\mu_\text{post}, \diag\bm\sigma_\text{post}^2)$ and a Gaussian with covariance $\mathbf I$ centered at the predicted mean, i.e., $\mathcal N(\bm\mu_\text{post}, \mathbf I)$. This is shown in Figure~\ref{fig:diagram}d, where each posterior is encouraged to stretch with covariance $\mathbf I$. Formally, the loss is 

\begin{small}
\begin{align}\nonumber
&J=J_\text{rec} +\lambda_\text{WAE}\cdot\widehat{\operatorname{MMD}}\\ 
&+ \lambda_\text{KL}\sum\nolimits_n \KL\left(\mathcal N(\bm\mu_\text{post}^{\n}, \diag(\bm\sigma_\text{post}^{\n})^2)\big\|\mathcal N(\bm\mu_\text{post}^{\n}, \mathbf I)\right)\label{eqn:obj}
\end{align}
\end{small}

\noindent\!While our approach appears heuristic, the next theorem shows its theoretical justification.

\begin{theorem}
Objective (\ref{eqn:obj}) is a relaxed optimization of the WAE loss (\ref{eqn:Jwae}) with a constraint on $\bm \sigma_\textnormal{post}$. (See Appendix~\ref{sec:B} for the proof.)\qed
\end{theorem}

We will show empirically that such auxiliary loss enables us to generate smoother and more diverse sentences in WAE. It, however, does not suffer from KL collapse as in VAEs. The auxiliary KL loss that we define for stochastic WAE is computed against a target distribution  $\mathcal N(\bm\mu_\text{post}^{\n}, \mathbf I)$ for each data sample $\mathbf x\n$. Here, the predicted posterior mean itself is used in the target distribution. As a result, this KL term does not force the model to learn the same posterior for all data samples (as in VAE), and thus, the decoder does not degrade to an unconditioned language model.

\section{Experiments} 

We evaluate WAE in sentence generation on the Stanford Natural Language Inference (SNLI) dataset~\cite{bowman2015large} as well as dialog response generation. All models use single-layer RNN with long short term memory (LSTM) units for both the encoder and decoder.  Appendix~\ref{sec:C} details our experimental settings.

\paragraph{VAE training.} VAE is notoriously difficult to train in the RNN setting. While different researchers have their own practice of training VAE, we follow our previous experience~\cite{VarAtt} and adopt the following tricks to stabilize the training: (1)  $\lambda_\text{VAE}$ was annealed in a $\operatorname{sigmoid}$ manner. We monitored the value of $\lambda \cdot \KL$ and stop annealing once it reached its peak value, known as \textit{peaking annealing}. (2) For word dropout, we started with no dropout, and gradually increased the dropout rate by $0.05$ every epoch until it reached a value of $0.5$. The effect of KL annealing is further analyzed in Appendix~\ref{sec:D}.

\begin{table}[!t] 
  \centering
  \resizebox{.475\textwidth}{!}{
    \begin{tabular}{|l||c||c|c|r|r|}
    \toprule
          & \textbf{BLEU}$^\uparrow$ & \textbf{PPL}$^\downarrow$  & \!\!\textbf{UniKL}$^\downarrow$ \!\! & \!\!\textbf{Entropy}\!\! & \textbf{AvgLen} \\
    \midrule
    \textbf{Corpus} & -     &    -   & -     &\!\!\!$\rightarrow$ 5.65  &\!\!\!$\rightarrow$ 9.6 \\
    \midrule\hline
    \textbf{DAE} & \textbf{86.35} &   146.2    & 0.178 & 6.23  & 11.0 \\
    \midrule
    \textbf{VAE} (KL-annealed) & 43.18 &  79.4  & 0.081 & 5.04  & 8.8 \\
    \midrule
    \textbf{WAE-D} $\lambda_{\text{WAE}}\!=\!3$ & 86.03 &   113.8    & 0.071 & 5.59  & 10.0 \\
    \textbf{WAE-D} $\lambda_\text{WAE}\!=\!10$& 84.29 &   104.9    & 0.073 & 5.57  & 9.9 \\
    \midrule
    \textbf{WAE-S}  $\lambda_{\text{KL}}=0.0$& 75.66 &    115.2  & 0.069 & \textbf{5.61}  & 9.9 \\
    \textbf{WAE-S} $\lambda_{\text{KL}}=0.01$& 82.01 &    84.9   & \textbf{0.058} & 5.26  & \textbf{9.4} \\
    \textbf{WAE-S} $\lambda_{\text{KL}}=0.1$& 47.63 &   \textbf{62.5}    & 0.150 & 4.65  & 8.7 \\
    %\textbf{WAE-S} $\lambda_\text{KL}=1.0$& 1 &  \textbf{5.57}   & 0.280 & 4.30  & 8.9 \\
    \bottomrule
    \end{tabular}%
    }
\caption{Results of SNLI-style sentence generation, where WAE is compared with DAE and VAE. \textbf{D} and \textbf{S} refer to the deterministic and stochastic encoders, respectively. $^{\uparrow/\downarrow}$The larger/lower, the better. For \textbf{Entropy} and \textbf{AvgLen}, the closer to corpus statistics, the better (indicated by the $\rightarrow$ arrow).}
\label{tab:overall}%
\end{table}%

\subsection{SNLI Generation} 

The SNLI sentences are written by crowd-sourcing human workers in an image captioning task. It is a massive corpus but with comparatively simple sentences (examples shown in Table~\ref{tab:samples}). This task could be thought of as domain-specific sentence generation, analogous to hand written digit generation in computer vision.

In Table~\ref{tab:overall}, we compare all methods  in two aspects. (1) We evaluate by \textbf{BLEU} \citep{papineni2002bleu} how an autoencoder preserves input information in a reconstruction task. (2) We also evaluate the quality of probabilistic sentence generation from the latent space. Although there is no probabilistic modeling of the latent space in DAE, we nevertheless draw samples from $\stdnorm$, which could serve as a non-informative prior. Perplexity (\textbf{PPL}) evaluates how fluent the generated sentences are. This is given by a third-party $n$-gram language model trained on the Wikipedia dataset. 
The unigram-KL (\textbf{UniKL}) evaluates if the word distribution of the generated sentences is close to that of the training corpus. Other surface metrics (entropy of the word distribution and average sentence length) also measure the similarity of the latent space generated sentence set to that of the corpus. 

We see that DAE achieves the best BLEU score, which is not surprising because DAE directly optimizes the maximum likelihood of data as a surrogate of word prediction accuracy. Consequently, DAE performs poorly for probabilistic sentence generation as indicated by the other metrics. 

VAE and WAE have additional penalties that depart from the goal of reconstruction. However, we see that WAEs, when trained with appropriate hyperparameters ($\lambda_{\text{WAE}},\lambda_{\text{KL}}$),  achieve close performance to DAE, outperforming VAE by 40 BLEU points. This is because VAE encodes each input's posterior to be close to the prior, from which it is impossible to perfectly reconstruct the data. 

Comparing the deterministic and stochastic encoders in WAE, we observe the same trade-off between reconstruction and sampling. However, our proposed stochastic encoder, with $\lambda_{\KL}=0.1$ for WAE, consistently outperforms VAE in the contradictory metrics BLEU and PPL. The hyperparameters $\lambda_\text{WAE}=10.0$ and $\lambda_{\KL}=0.01$ appear to have the best balance between reconstruction, sentence fluency, as well as similarity to the original corpus. 

Moreover, all our WAEs are trained without annealing or word dropout. It is significantly simpler than training a VAE, whose KL annealing typically involves a number of engineering tricks, such as the time step when KL is included, the slope of annealing, and the stopping criterion for annealing.

\subsection{Dialog Generation} We extend WAE to an encoder-decoder framework (denoted by WED) and evaluate it on the DailyDialog corpus~\cite{li2017dailydialog}.\footnote{In our pilot experiment, we obtained a BLEU-4 score of 6 by training a pure Seq2Seq model with LSTM units for 200 epochs, whereas \newcite{li2017dailydialog} report 0.009 BLEU-4 and \newcite{2.8} report 2.84 BLEU-4. Due to our unreasonably high performance, we investigated this in depth and found that the training and test sets of the DailyDialog corpus have overlaps. For the results reported in our paper, we have removed duplicate data in the test set, which is also available on our website~(Footnote 1). To the best of our knowledge, we are the first to figure out the problem, which, unfortunately, makes comparison with previous work impossible.} We follow~\newcite{VarAtt}, using the encoder to capture an utterance and the decoder to generate a reply.

Table~\ref{tab:dialog} shows that WED with a deterministic encoder (WED-D) is better than the variational encoder-decoder (VED) in BLEU scores, but the generated sentences lack variety, which is measured by output entropy and the percentage of distinct unigrams and bigrams \cite[Dist-1/Dist-2,][]{diversity-promoting}, evaluated on the generated test set responses.

We then applied our stochastic encoder for WED and see that, equipped with our KL-penalized stochastic encoder, WED-S outperforms DED, VED, and WED-D in all diversity measures. WED-S also outperforms VED in generation quality, consistent with the results in Table~\ref{tab:overall}.

\begin{table}[!t]
  \centering
  \resizebox{.475\textwidth}{!}{
    \begin{tabular}{|l|c|c||c|c|c|}
    \toprule
          & \textbf{BLEU-2} & \textbf{BLEU-4} & \textbf{Entropy} & \textbf{Dist-1} & \textbf{Dist-2} \\
    \midrule\hline
    \textbf{Test Set} & \textbf{-} & \textbf{-} & 6.15  & 0.077 & 0.414 \\
    \midrule
    \textbf{DED} &  3.96 & 0.85  & 5.55  & 0.044 & 0.275 \\
    \textbf{VED} & 3.26  & 0.59  & 5.45  & 0.053 & 0.204 \\
    \textbf{WED-D} & \textbf{4.05} & \textbf{0.98} & 5.53  & 0.042 & 0.272 \\
    \textbf{WED-S} & 3.72  & 0.69  & \textbf{5.59} & \textbf{0.066} & \textbf{0.309} \\
    \bottomrule
    \end{tabular}%
    }
    \caption{Results on dialog generation, where VED/WED hyperparameters for each model were chosen by Table~\ref{tab:overall}.}
  \label{tab:dialog}
\end{table}%

\section{Conclusion} 

In this paper, we address the difficulty of training VAE by using a Wasserstein autoencoder (WAE) for probabilistic sentence generation. WAE implementation can be carried out with either a deterministic encoder or a stochastic one. The deterministic version achieves high reconstruction performance, but lacks diversity for generation. The stochastic encoder in the original form may collapse to a Dirac delta function, shown by both a theorem and empirical results. We thus propose to encourage stochasticity by penalizing a heuristic KL loss for WAE, which turns out to be a relaxed optimization of the Wasserstein distance with a constraint on the posterior family.

We evaluated our model on both SNLI sentence generation and dialog systems. We see that WAE  achieves high reconstruction performance as DAE, while retaining the probabilistic property as VAE. Our KL-penalty further improves the stochasticity of WAE, as we achieve the highest performance in all diversity measures.

\section*{Acknowledgments}
We would like to acknowledge Yiping Song and Zhiliang Tian for their independent investigation on the DailyDialog corpus. We also thank Yanran Li, one of the authors who released DailyDialog, for discussion on this issue. This work was supported in part by the NSERC grant RGPIN-261439-2013 and an Amazon Research Award.
\bibliography{acl2018}
\bibliographystyle{acl_natbib}

\appendix
\section{Proof of Theorem 1}\label{sec:A}
\noindent\textbf{Theorem 1. } \textit{Suppose we have a Gaussian family $\mathcal{N}(\bm\mu,\diag\bm\sigma^2)$, where  $\bm \mu$ and $\bm \sigma$ are parameters. The covariance is diagonal, meaning that the variables are independent. If the gradient of $\bm \sigma$ completely comes from sample gradient and $\bm \sigma$ is small at the beginning of training, then the Gaussian converges to a Dirac delta function with stochastic gradient descent, i.e., $\bm \sigma\rightarrow\bm 0$.}
	
\medskip
\noindent\textit{Proof.} For the predicted posterior $\mathcal{N}(\bm \mu,\diag\bm \sigma^2)$ where all dimensions are independent, we consider a certain dimension, where the sample is $z_i\sim \mathcal{N}(\mu_i,\sigma_i^2)$.

We denote the gradient of $J$ wrt to $z_i$ at $z_i=\mu_i$ by $g_i\overset{\Delta}=\frac{\partial J}{\partial z_i}\big|_{z_i=\mu_i}$. 
At a particular sample $z_i\jj$ around $\mu_i$, the gradient $g_i\jj$ is
\begin{align}
g_i\jj &\overset{\Delta}{=}\frac{\partial J}{\partial z_i}\bigg |_{z_i=z_i\jj}\\ \label{eqn:taylor}
&\approx g_i+\frac{\partial^2 J}{\partial \mu_i^2}(z_i\jj-\mu_i)\\
&\overset{\Delta}=g_i+k(z_i\jj-\mu_i)\\
&=g_i+k(\mu_i+\epsilon\jj\sigma_i-\mu_i)\\
&=g_i+k\sigma_i\epsilon\jj
\end{align}
where (\ref{eqn:taylor}) is due to Taylor series approximation, if we assume $ \sigma_i^2$ is small and thus $z_i\jj$ is near $\mu_i$. $k$ denotes  $\frac{\partial^2 J}{\partial \mu_i^2}$.

We compute the expected gradient wrt to $\sigma_i$ for $\epsilon\sim\mathcal{N}(0,1)$. The assumption of this theorem is that the gradient of $\mu_i$ and $\sigma_i$ completely comes from the sample $z_i$. By the chain rule, we have

\begin{align}
&\E_{\epsilon\jj\sim\mathcal N(0,1)}\left[\frac{\partial J}{\partial \sigma_i}\right]\\
=&\E_{\epsilon\jj\sim\mathcal{N}(0,1)}\left[
\frac{\partial J}{\partial z_i\jj}\cdot 
\frac{\partial z_i\jj}{\partial \sigma_i}
\right]\\
=&\E_{\epsilon\jj\sim\mathcal{N}(0,1)}\left[
(g_i+k\sigma_i\epsilon\jj)\cdot{\epsilon\jj}
\right]\\
=&\E_{\epsilon\jj\sim\mathcal N(0,1)}\left[g_i\epsilon_i\jj+k\sigma_i(\epsilon\jj)^2\right]\\
=& k\sigma_i
\end{align}
Notice that $k>0$ if we are near a local optimum (locally convex).

In other words, the expected gradient of $\sigma_i$ is proportional to $\sigma_i$. According to stochastic gradient descent (SGD), $\sigma_i$ will converge to zero.\qed 

\begin{figure}[!t]
\centering
\includegraphics[width=\linewidth]{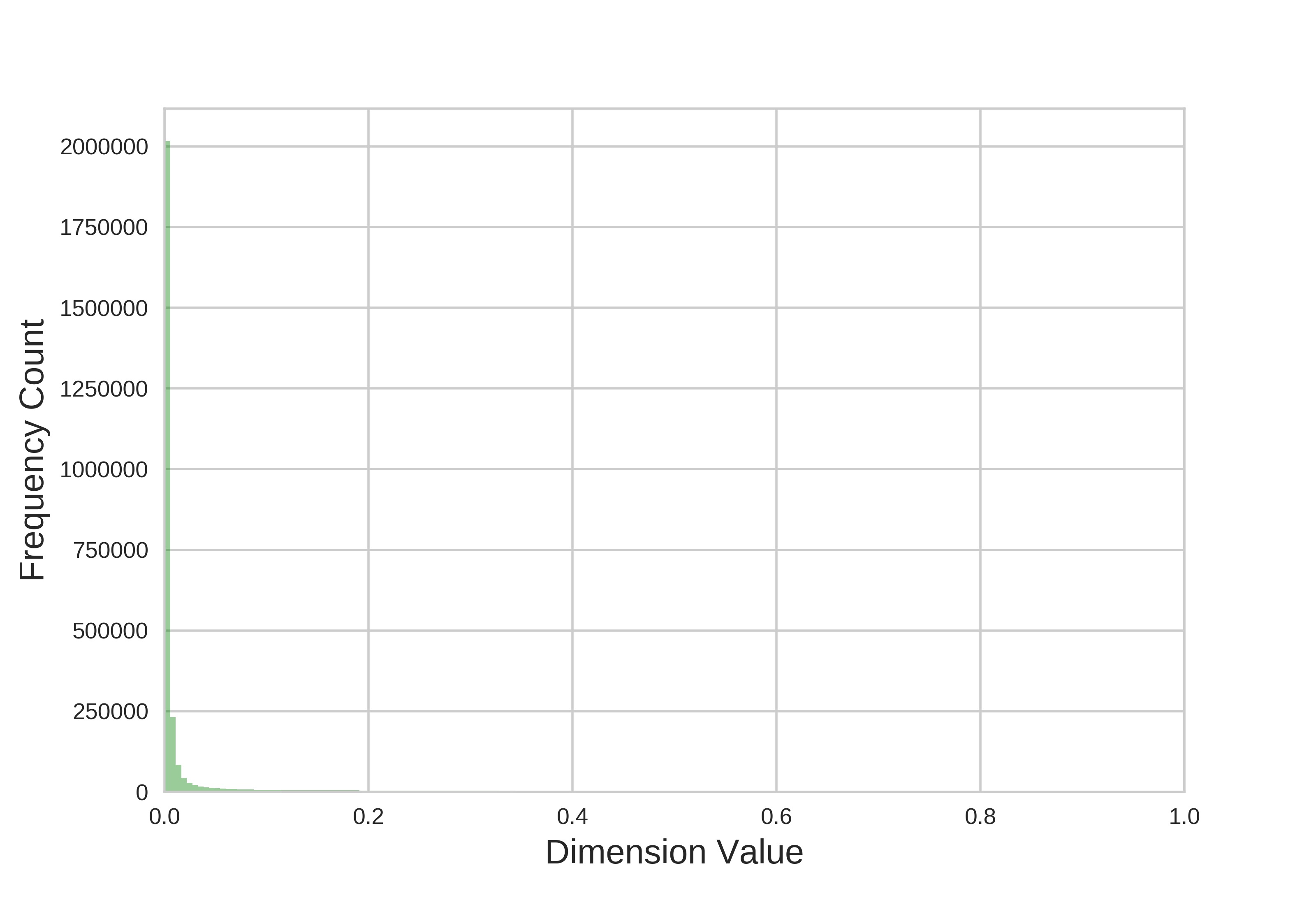}
(a) $\lambda_{\KL}=0$
\includegraphics[width=\linewidth]{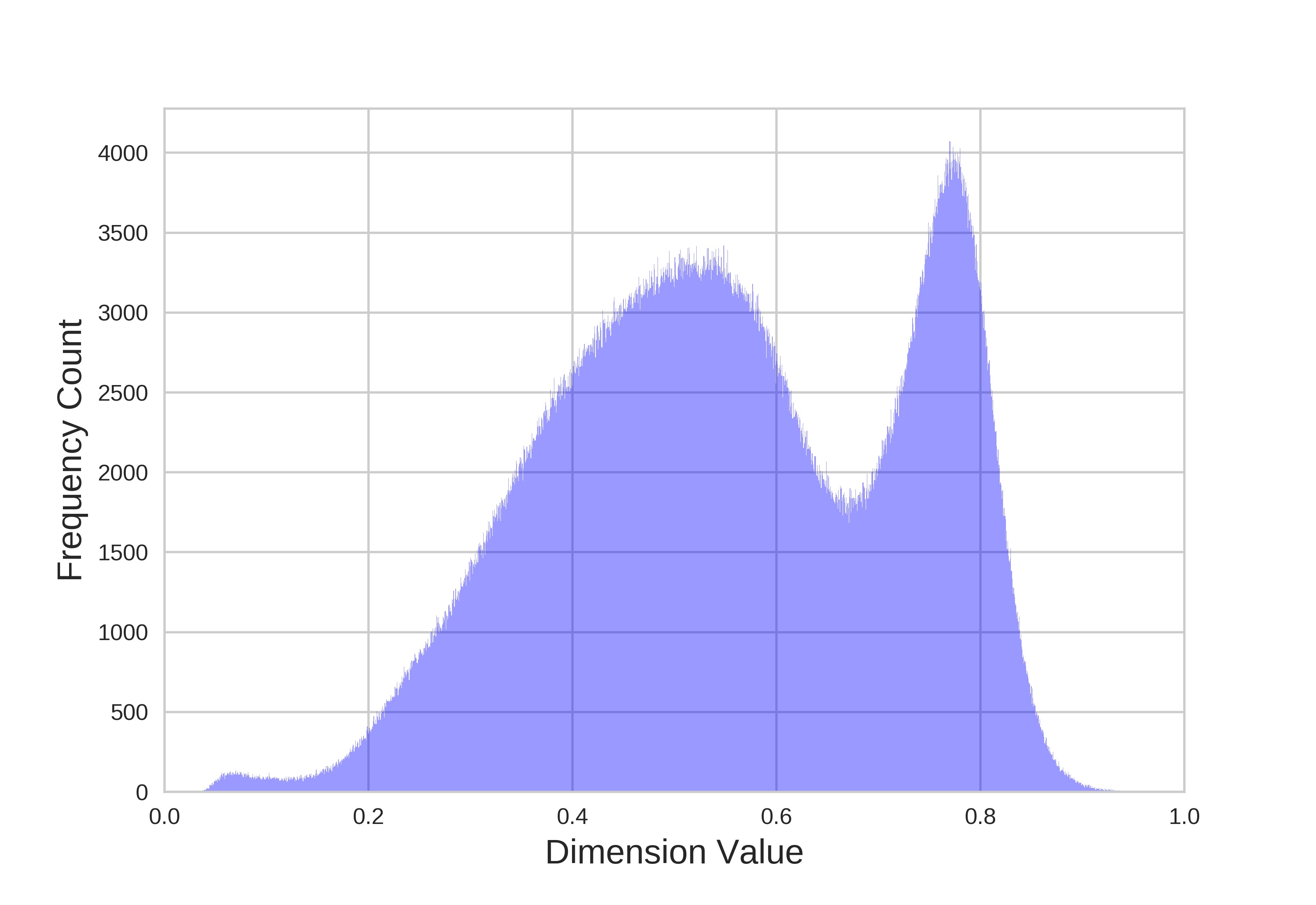}
(b) $\lambda_{\KL}=0.01$
\includegraphics[width=\linewidth]{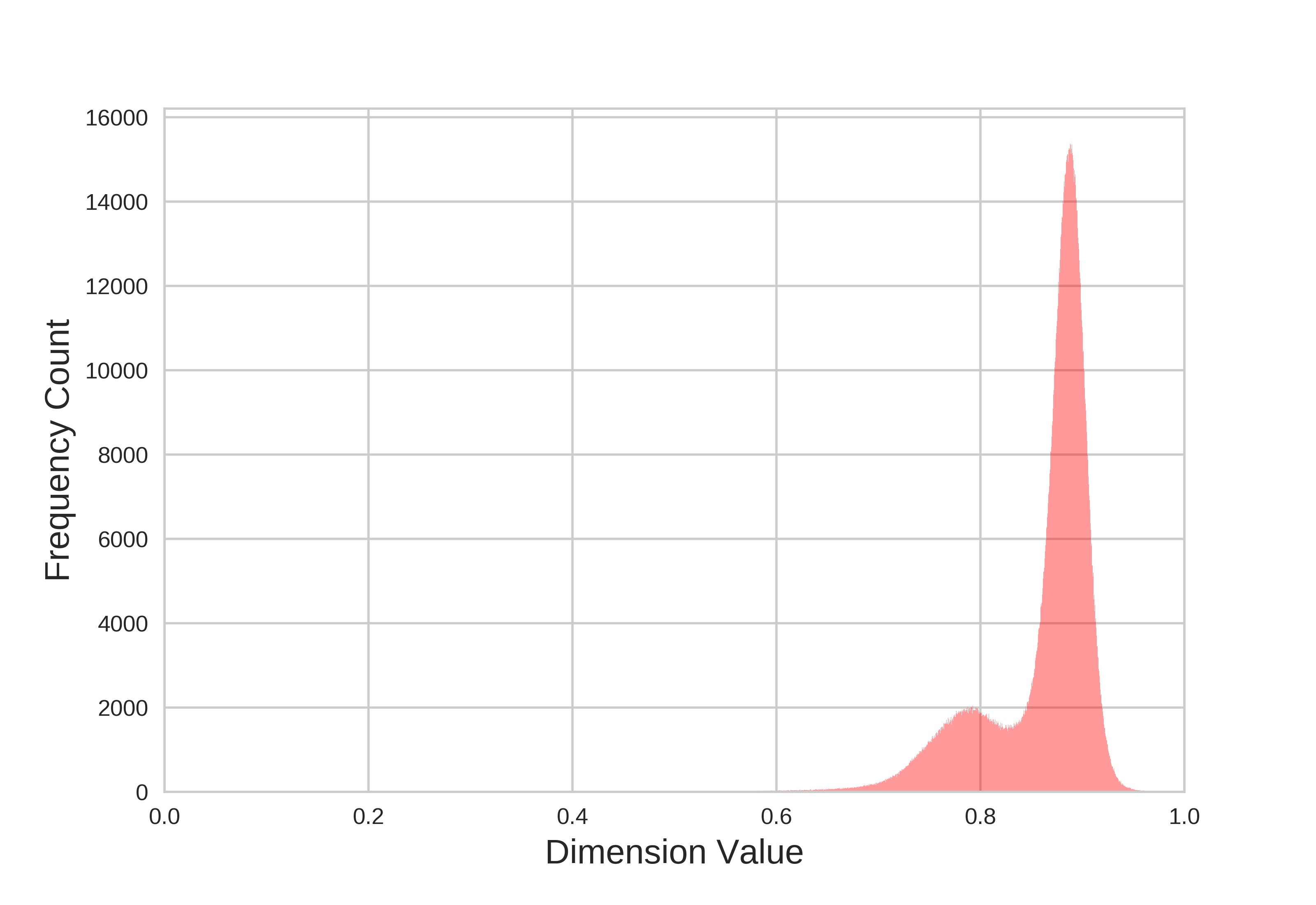}
(c) $\lambda_{\KL}=0.1$
\caption{The histograms of $\sigma$ in the posterior of the WAE in the SNLI experiment. In the plot, there are 200 buckets in the range of $(0,1)$.}\label{fig:waeKL}
\end{figure}

\bigskip
The theorem assumes $\bm \sigma^2$ is small, compared with how $J$ changes in the latent space. In practice, the encoded vectors of different samples may vary a lot, whereas if we sample different vectors from a certain predicted multi-variate Gaussian, we would generally obtain the same sentence. Therefore, $J$ is kind of smooth in the latent space. The phenomenon can also be verified empirically by plotting the histogram of $\sigma$ in WAE with a stochastic Gaussian encoder  (Figure~\ref{fig:waeKL}).  We see that if the KL coefficient $\lambda_{\KL}$ is 0, meaning that the gradient of $\sigma$ comes only from the samples, then most $\sigma$'s collapse to~0.

Notice, however, that the theorem does not suggest a stochastic WAE and a deterministic WAE will yield exactly the same result, as their trajectories may be different.

\section{Proof of Theorem 2}\label{sec:B}

\textbf{Theorem 2.} \textit{Objective (\ref{eqn:obj}) is a relaxed optimization of the WAE loss (\ref{eqn:Jwae}) with a constraint on $\bm \sigma_\textnormal{post}$.}

\medskip
\noindent\textit{Proof.} Objective (\ref{eqn:obj}) optimizes 

\begin{small}
\begin{align}\nonumber
&J=J_\text{rec} +\lambda_\text{WAE}\cdot\widehat{\operatorname{MMD}}\\ 
&+ \lambda_\text{KL}\sum\nolimits_n \KL\left(\mathcal N(\bm\mu_\text{post}^{\n}, \diag(\bm\sigma_\text{post}^{\n})^2)\big\|\mathcal N(\bm\mu_\text{post}^{\n}, \mathbf I)\right)\nonumber
\end{align}
\end{small}

The first two terms are the WAE loss, whereas the last penalty relaxes the following optimization problem

\begin{small}\begin{align}\nonumber
&\text{minimize}\ \  J_\text{rec} + \lambda_\text{WAE}\cdot\widehat{\operatorname{MMD}}\\ \nonumber
&\text{subject to}\\   
&\ \ \sum\nolimits_n \KL\left(\mathcal N(\bm\mu_\text{post}^{\n}, \diag(\bm\sigma_\text{post}^{\n}{}^2))\big\|\mathcal N(\bm\mu_\text{post}^{\n}, \mathbf I)\right) < C\label{eqn:constraint}
\end{align}
\end{small}

\vspace{-.5cm}
\noindent for some constant $C$.

As known, the KL divergence between two (univariant) Gaussian distributions is
\begin{align}\nonumber
&\KL(\mathcal{N}(\mu_1, \sigma_1)\|(\mathcal{N}(\mu_2, \sigma_2)))\\
=&\log\frac{\sigma_2}{\sigma_1}+\frac{\sigma_1^2+(\mu_1-\mu_2)^2}{2\sigma_2^2}-\frac12
\end{align}

The constraint in (\ref{eqn:constraint}) is equivalent to 
\begin{equation}
\sum_n\sum_i\left[ -\log \sigma_i\n + \dfrac12 (\sigma_i\n)^2\right]< C
\end{equation}

In other words, our seemingly heuristic KL penalty optimizes the Wasserstein loss, while restricting the posterior family.\qed

\section{Implementation Details}\label{sec:C}
All models were trained with the Adam optimizer \cite{kingma2014adam} with $\beta_1=0.9$ and $\beta_2=0.999$. In all our experiments, we feed the sampled latent vector $\bm z$ to each time step of the decoder. Task-specific settings are listed in Table~\ref{tab:settings}. 

\begin{table}[!t]
  \centering
  SNLI Experiment
  \resizebox{.475\textwidth}{!}{
    \begin{tabular}{|p{5cm}|p{6cm}|}
    \toprule
    \textbf{LSTM Hidden Dimension} & 100d, single layer \\
    \midrule
    \textbf{Word Embeddings} & 300d, pretrained on SNLI Corpus \\
    \midrule
    \textbf{Latent Dimension} & 100d \\
    \midrule
    \textbf{Epochs} & 20 \\
    \midrule
    \textbf{Learning Rate} & Fixed rate of 0.001 \\
    \midrule
    \textbf{Batch Size } & 128 \\
    \midrule
    \textbf{Max Sequence Length} & 20 \\
    \midrule
    \textbf{Vocab Size} & 30000 \\
    \bottomrule
    \end{tabular}%
    }
    
\bigskip
  \centering
  Dialog Experiment
  \resizebox{.475\textwidth}{!}{
    \begin{tabular}{|p{5cm}|p{6cm}|}
    \toprule
    \textbf{LSTM Hidden Dimension} & 500d, single layer \\
    \midrule
    \textbf{Word Embeddings} & 300d, pretrained on DialyDialog Corpus \\
    \midrule
    \textbf{Latent Dimension} & 300d \\
    \midrule
    \textbf{Epochs} & 200 \\
    \midrule
    \textbf{Learning Rate} & Initial rate of 0.001, multiplicative decay of 0.98 until a minimum of 0.00001 \\
    \midrule
    \textbf{Batch Size } & 128 \\
    \midrule
    \textbf{Max Sequence Length} & 20 \\
    \midrule
    \textbf{Vocab Size} & 20000 \\
    \bottomrule
    \end{tabular}%
    }\caption{Experimental settings.}\label{tab:settings}
\end{table}%

\section{VAE Training Difficulties}\label{sec:D}
\begin{figure}[!t]
\centering
~\hspace{-.3cm}\includegraphics[width=1.05\linewidth]{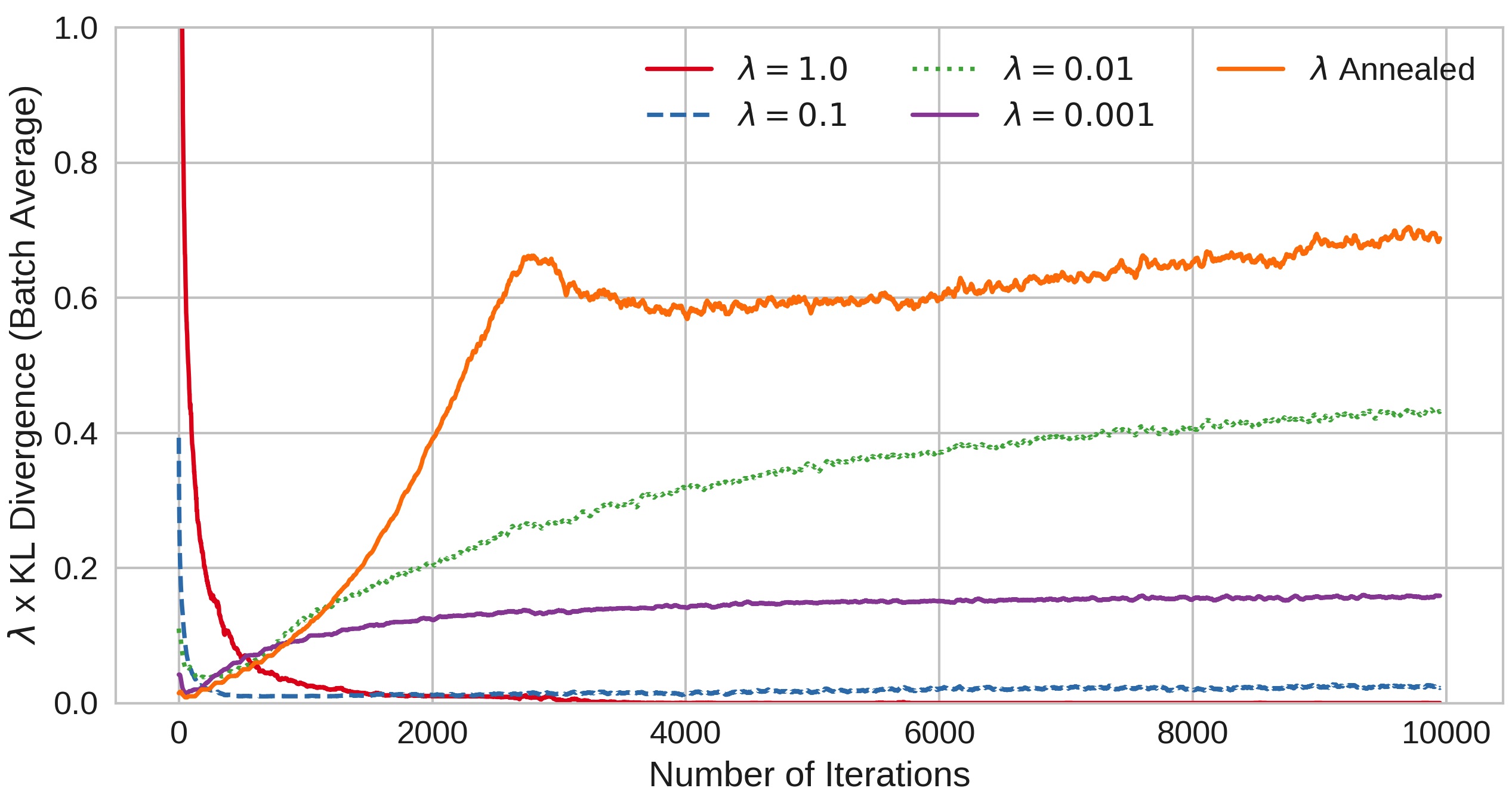}
\caption{Learning curves of the KL term in the VAE loss function ($\lambda \cdot \KL$) for different values of $\lambda$, and the variant where $\lambda$ is annealed.}\label{fig:KL}
\end{figure}

It is a common practice that training VAEs involves KL annealing and word dropout, which further consists of hacks for tuning hyperparameters.
We conducted an experiment of training VAE without KL annealing. In Figure~\ref{fig:KL}, we present the KL loss (weighted by $\lambda_\text{VAE}$) during the training process for different values of $\lambda_\text{VAE}$. The KL loss is believed to be an important diagnostic measure to indicate if the latent space is ``variational''~\cite{yang2017improved, higgins2017beta, burgess2018understanding}. We see that if the penalty is too large, KL simply collapses to zero ignoring the entire input, in the case of which, the decoder becomes an unconditioned language model. On the other hand, if the KL penalty is too small, the model tends to become more \textit{deterministic} and the KL term does not play a role in the training. This is expected since in the  limit of $\lambda_{\text{VAE}}$ to $0$, the model would probably ignore the KL term and becomes a deterministic autoencoder (shown also by Theorem~1). 

The VAE with collapsed KL does not exhibit interesting properties such as random sampling for probabilistic sequence generation \citep{bowman2015generating}.
As seen in Table~\ref{tab:samples}, the generated sentences by VAE without annealing are very close to each other. This is because VAE's encoder does not capture useful information in the latent space, which is simply ignored during the decoding phase. By sampling the latent space, we do not obtain varying sentences. The empirical evidence verifies our intuition.

\begin{table}[!t]
  \centering
  \resizebox{\linewidth}{!}{ 
    \begin{tabular}{|l|}
    \toprule
    \textbf{Training Samples} \\
    \midrule
    \midrule
    \textit{a mother and her child are outdoors.} \\
    \textit{the people are opening presents.} \\
    \textit{the girls are looking toward the water.} \\
    \textit{a small boy walks down a wooden path in the woods.} \\
    \textit{a person in a green jacket it surfing while holding on to a line.} \\
    \midrule
    \textbf{DAE} \\
    \midrule
    \midrule
    \textit{two families walking in a towel down alaska sands a cot .} \\
    \textit{a blade is rolling its nose furiously paper .} \\
    \textit{a woman in blue shirts is passing by a some beach} \\
    \textit{transporting his child are wearing overalls .} \\
    \textit{a guys are blowing on professional thinks the horse .} \\
    \midrule
        \textbf{VAE without Annealing} \\
    \midrule
    \midrule
    \textit{a man is playing a guitar .} \\
    \textit{a man is playing with a dog .} \\
    \textit{a man is playing with a dog .} \\
    \textit{a man is playing a guitar .} \\
    \textit{a man is playing with a dog .} \\
    \midrule
    \textbf{VAE with Annealing} \\
    \midrule
    \midrule
    \textit{the band is sitting on the main street .} \\
    \textit{couple dance on stage in a crowded room .} \\
    \textit{two people run alone in an empty field .} \\
    \textit{the group of people have gathered in a picture .} \\
    \textit{a cruise ship is docking a boat ship .} \\
    \midrule
    \textbf{VAE vMF} ($\kappa$ fixed)\\
    \midrule\midrule
    \textit{a car is a and and a blue shirt top is .}\\
\textit{two children are playing on the group in are the the . the}\\
\textit{a child and a adult and}\\
\textit{the young is playing for a picture a are playing to}\\
\textit{a little is playing a background . .}\\
    \midrule
    \midrule
    \textbf{WAE-D} ($\lambda_\text{WAE}=10$)\\
    \midrule
    \midrule
    \textit{the lone man is working .} \\
    \textit{the group of men is using ice at the sunset .} \\
    \textit{a family is outside in the background .} \\
    \textit{two women are standing on a busy street outside a fair} \\
    \textit{a tourists is having fun on a sunny day} \\
    \midrule
    \textbf{WAE-S} ($\lambda_\text{WAE}=10$, $\lambda_\text{KL}=0.01$) \\
    \midrule
    \midrule
    \textit{an asian man is dancing in a highland house .} \\
    \textit{a person wearing a purple snowsuit jumps over the tree .} \\
    \textit{the vocalist is at the music and dancing with a microphone .} \\
   \textit{a young man is dressed in a white shirt cleaning clothes .} \\
    \textit{three children lie together and a woman falls in a plane .} \\
    \bottomrule
    \end{tabular}%
    }
      \caption{Sentences generated by randomly sampling from the prior  for different models.}
  \label{tab:samples}
\end{table}%

A recent study~\cite{spherical} propose to get rid of KL annealing by using the von Mises--Fisher (vMF) family of posterior and prior. In particular, they set the prior to the uniform distribution on a unit hypersphere, whereas the posterior family is normal distribution on the surface of the same sphere. They fix the standard deviation (parametrized by $\kappa$) of the posterior, so that their KL is a constant and annealing is not required. This, unfortunately, loses the privilege of learning uncertainty in the probabilistic modeling. Examples in Table~\ref{tab:samples} show that, while we have reproduced the reconstruction negative log-likelihood with vMF-VAE (the metric used in their paper), the generated sentences are of poor quality. As also suggested by \newcite{spherical}, if the posterior uncertainty in vMF is made learnable, it re-introduces the KL collapse problem, in which case, the KL annealing is still needed.

By contrast, WAEs for sequence-to-sequence models are trained without any additional optimization strategies such as annealing. Even in our stochastic encoder, the KL penalty does not make WAE an unconditioned language model, because it does not force the encoded posterior to be the same for different input sentences.

\section{Qualitative Samples}\label{ref:E}
Table~\ref{tab:samples} shows sentences generated by randomly sampling points in the latent space for different models, along with sample sentences from the training set. 
They provide a qualitative understanding of each model's performance.

\end{document}